\documentclass[letterpaper, 10 pt, conference]{ieeeconf}  
\IEEEoverridecommandlockouts                              

\usepackage{microtype}
\usepackage{comment}

\usepackage{graphicx} 
\graphicspath{{./figures/}}
\usepackage{subcaption}
\usepackage{float}

\usepackage{amsmath} 
\usepackage{amssymb}  
\usepackage{units}
\usepackage{bm} 
\usepackage{multirow}

\usepackage{algorithm}
\usepackage{algpseudocode}

\makeatletter
\let\NAT@parse\undefined
\makeatother

\usepackage[square, numbers, sort&compress]{natbib}
\usepackage[hidelinks]{hyperref}
\usepackage{cleveref}

\usepackage{graphicx} 
\graphicspath{{./figures/}}
\usepackage{subcaption}
\usepackage{float}
\usepackage[font=small]{caption}
\usepackage{xparse}
\usepackage{xcolor}
\usepackage{stfloats}

\title{\LARGE \bf Learning Task Space Actions for Bipedal Locomotion}

\author{Helei Duan$^1$, Jeremy Dao$^1$, Kevin Green$^1$, Taylor Apgar$^2$, Alan Fern$^1$, Jonathan Hurst$^1$
\thanks{*This work is supported by the NSF Grant No. IIS-1849343, DGE-1314109, and DARPA Contract W911NF-16-1-0002.}
\thanks{$^1$Collaborative Robotics and Intelligent Systems Institute, Oregon State University, Corvallis, Oregon, 97331, USA. Email: \{{\tt\footnotesize duanh, daoje, greenkev, afern, jhurst@oregonstate.edu}\}. }
\thanks{$^2$Agility Robotics, Albany, Oregon, 97321, USA.
}
}

\begin{document}

\maketitle
\thispagestyle{empty}
\pagestyle{empty}

\begin{abstract}
Recent work has demonstrated the success of reinforcement learning (RL) for training bipedal locomotion policies for real robots. 
This prior work, however, has focused on learning joint-coordination controllers based on an objective of following joint trajectories produced by already available controllers. 
As such, it is difficult to train these approaches to achieve higher-level goals of legged locomotion, such as simply specifying the desired end-effector foot movement or ground reaction forces. 
In this work, we propose an approach for integrating knowledge of the robot system into RL to allow for learning at the level of task space actions in terms of feet setpoints. 
In particular, we integrate learning a task space policy with a model-based inverse dynamics controller, which translates task space actions into joint-level controls.
With this natural action space for learning locomotion, the approach is more sample efficient and produces desired task space dynamics compared to learning purely joint space actions. 
We demonstrate the approach in simulation and also show that the learned policies are able to transfer to the real bipedal robot Cassie.
This result encourages further research towards incorporating bipedal control techniques into the structure of the learning process to enable dynamic behaviors.
\end{abstract}

\section{INTRODUCTION} 

Robotic legged locomotion has benefited from decades of focused analysis and rigorous control theory.
Many of the most successful and capable systems choose to control actions in task space by controlling foot placements, ground reactions forces (GRFs), swing foot motions, or center-of-mass (CoM) motion to maintain balance and stability. 
In the simplest case, reduced order models often parameterize locomotion dynamics by choosing a particular set of actions in task space, such as feet setpoints relative to CoM along with the controlled impedance, so that the model can generate compliant behaviors and reproduce the characteristic dynamics observed from animals \cite{Green2020, Geyer2006, Blum2014}.

\begin{figure}
    \centering
    \includegraphics[width=0.48\textwidth]{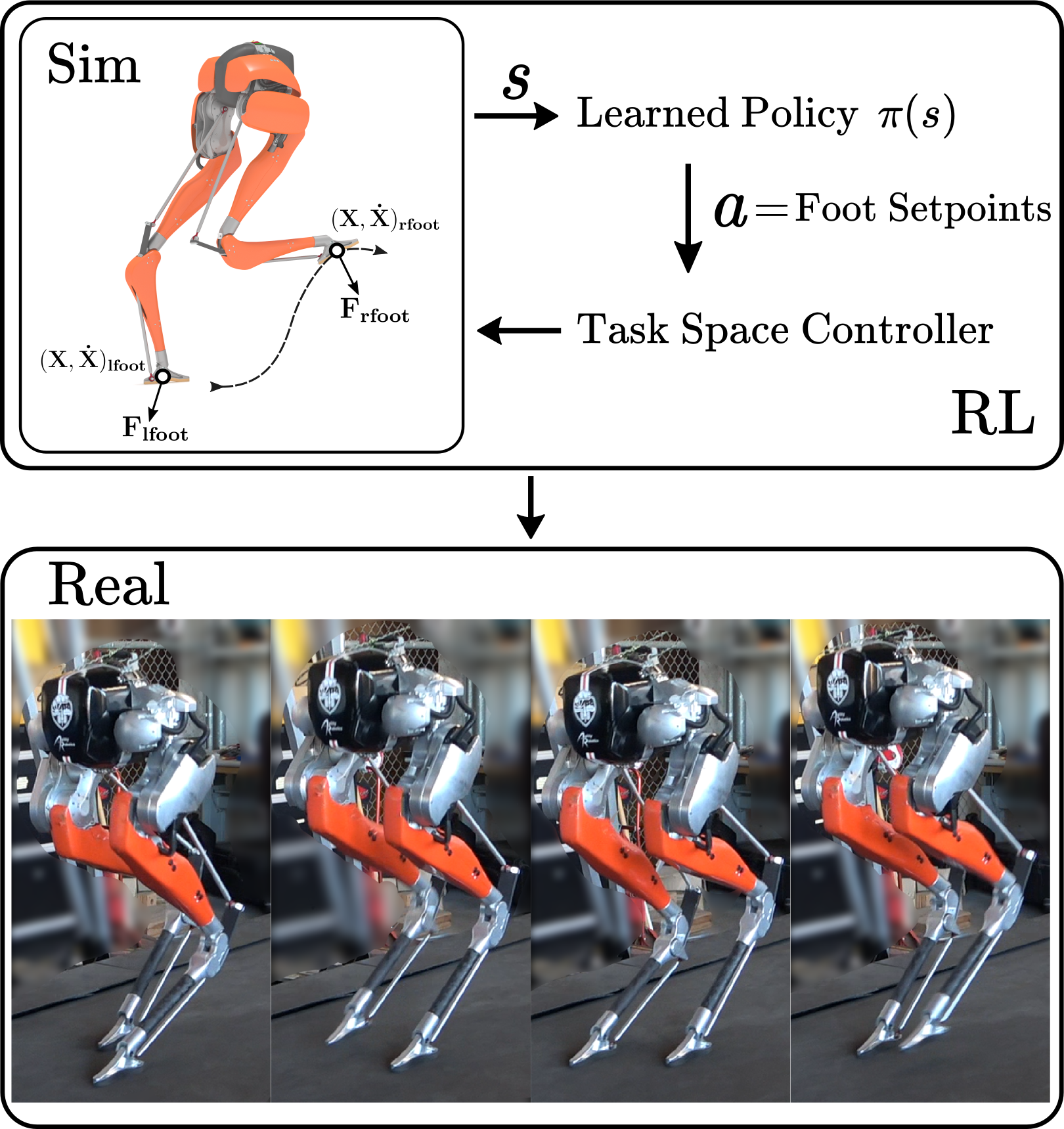}
    \caption{
    Learned policy outputs actions represented in task space to directly control $\mathbf{X, \dot X, F}$. The learned policy produces the desired task space dynamics in simulation and on hardware Cassie. 
    }
    \label{fig:title}
\end{figure}

By contrast, current model-free reinforcement learning (RL) methods for controlling legged robots mostly focus on learning joint control actions and defining the reward to track joint space trajectories \cite{Xie2019, Tan-RSS-18, Siekmann-RSS-20}. 
Representing learned actions in joint space may obstruct the policy from directly optimizing primary control interests and regulating task space dynamics. 
Joint space policies must implicitly re-learn significant information about the relation between task space and joint space in order to coordinate the joints for successful control of task space behaviors. 
These approaches also lack an interface to many successful model-based controllers, which exploit the task space formulation of the locomotion problems. 


Our approach aims to address such drawbacks by learning at the level of task space for legged locomotion.  
Actions in task space, such as feet movements, can directly influence the primary control interests and the behaviors of the agents, leading to more effective exploration than noisy, undirected exploration in joint space.
Furthermore, the known models often come with the use of model-based controllers that maps from task space dynamics to joint-level commands.
This allows the learned policies to interface with the same language as existing reduced order models \cite{Green2020} and whole-body control methods \cite{Apgar2018a, Luo2018a}.

In our proposed method, the learned policy outputs actions directly in task space as residual setpoints of the feet relative to the robot base. In particular, we integrate inverse dynamics reasoning about the known robot model so that the RL-learned task space actions are converted to joint commands. Thus, the agent does not need to learn about joint coordination and can explore directly in the more meaningful task space. In addition, the policy can now directly learn from reward functions that align with task space interests.  

Our first contribution is to show that this integrated approach can significantly reduce the sample requirements of reinforcement learning on bipedal locomotion, which agrees with other task space formulation in non-locomotion domains \cite{Martin-Martin2019, Varin2019a}. 
Our second contribution is to demonstrate the successful steady-state walking in simulation and on hardware using the bipedal robot Cassie. We use spring mass models to generate a library of desired task space dynamics as the references for the learning process. By integrating learning with the task space controller, the learned control policies can adapt the references and produce desired control interests. To our knowledge this is the first demonstration of a task space learned policy for real-world dynamic bipedal locomotion. 

\section{RELATED WORK} 

\begin{figure*}[t]
    \centering
    \vspace{0.3cm}
    \includegraphics[width=0.98\textwidth]{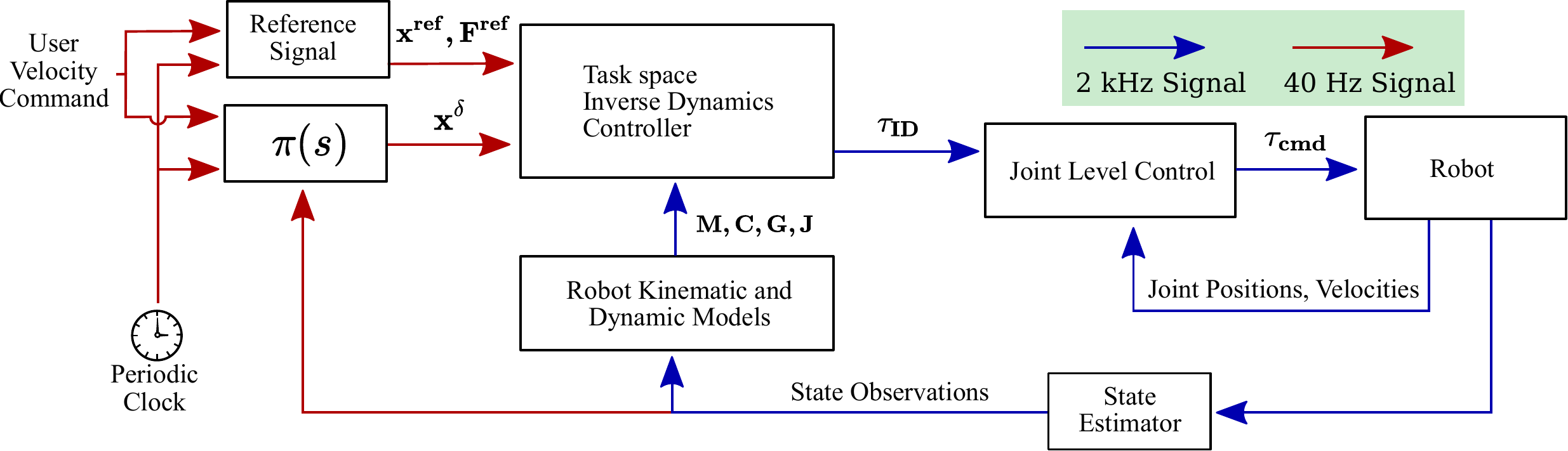}
    \caption{Approach Overview: 
    The output of the neural network policies $\pi(s)$ are residual setpoints of the feet relative to the robot base. 
    These residual setpoint components $\mathbf{x^{\delta} \in \mathbb{R}^6}$ together with the reference signals $\mathbf{ \{x^{ref}, F^{ref}\}}$ form the overall commands (red colored, updated at 40Hz) to the inverse dynamics control of the feet relative to the base. 
    Furthermore, learned actions are used in different control strategies depending on the gait phase. 
    For stance foot, the learned actions will be composed with task space impedance to produce forces that are added to the feedforward force to stabilize the robot.
    For swing foot, the learned actions will be converted to desired foot accelerations by task space PD control to then create compliant swing motions. 
    Signals from the policy are commands to the inverse dynamics controller which maps commanded setpoints to torques (blue colored, 2kHz). Details are in Section \ref{sec:control}. }
    \label{fig:framework}
    \vspace{-.5cm}
\end{figure*} 
\textbf{Action Space of RL for Dynamic Robot Control.}
In RL problems, the action space defines the range of options agents can take to influence dynamics and make transitions between states. 
When using RL to generate control policies for dynamic robots, such as legged robots, the action space also defines the direct interface from policy to low-level actuator commands, such as joint positions or torques. 
Previous research \cite{Peng2017a} showed that using joint setpoints as the policy output into a joint PD controller provides a better interface to control fast locomotion dynamics than pure torques or joint velocities. 
Joint position control has then been used extensively in RL approaches for controlling legged robots in simulation and on hardware \cite{Xie2019, Hwangbo2019}. 

Recent studies \cite{Martin-Martin2019, Varin2019a, 8793506} proposed a set of new action spaces for contact-sensitive manipulation control policies by defining the action space in the manipulator's end-effector space (task space). 
These approaches show that, by leveraging the kinematic and dynamic models of the robot, the learning process is more sample efficient than having actions defined in joint space. 
The learned policy is also able to better perform contact-rich tasks with constrained dynamics on redundant and fixed-based manipulators. 
In our work, we extend this idea and show the sample efficient learning even when the learning method is applied to underactuated floating-base robots where contact state changes more frequently and is fundamental for system stability.  

\textbf{Model-based Control for Bipedal Robots.}
Many successful model-based controllers for bipedal robots center on task space control methods to produce desired CoM and feet dynamics. These methods are integrated closely with the known robot models that are readily available. 
For example, \cite{Apgar2018a} uses an operational space controller to generate robust gaits while the dynamics model of the robot is used to perform fast online optimization. 
\cite{Martin2017a} proposes tracking the CoM trajectory and GRFs with LQR and an inverse dynamics controller. 
Despite many successes with model-based control methods, less is known on how to create a structured approach for learning locomotion policies that can work with existing model-based control on physical hardware. 
Our approach aims to provide an integrated method by bridging both sides through the same language defined in task space. 

\textbf{RL with Model-based Controllers.} 
Combining conventional feedback control with reinforcement learning can be done in various ways to improve sample efficiency and take advantage of existing controllers. 
By decomposing actuator commands into both learned- and model-based controllers, the residual learning approach \cite{Johannink2019} has shown promise towards merging both sides. The method used in \cite{9197125} improves sample efficiency by further dividing the control space in terms of uncertainty of the controller. 
Learning only happens near the uncertain regions where the model-based controller has less confidence. 
\cite{Xie2020} proposed using RL to generate foot contact sequences that are sent to a low-level task space controller on a quadruped, which demonstrates an integrated way of generating control policies for quadrupeds using a linearized centroidal dynamics model. 
Our aim is to learn reactive policies in task space for a human scale bipedal robot. 
We formulate the low-level task space controller into swing and stance phases that takes commands from learned policies.

\section{Control Structure}
\label{sec:control}

To facilitate the policy directly learning locomotion behaviors in task space, this section explains the task space control structure to which the learned controller issues commands. The overall control diagram is illustrated in Fig \ref{fig:framework}.

\subsection{Task Space Constrained Dynamics}
The Cassie robot's dynamics can be written in terms of its generalized, floating base coordinates. 
Each leg contains two passive springs and a closed kinematic loop.
Following \cite{Apgar2018a}, we treat all of the passive spring joints as rigid for the purposes of control. 
We can then eliminate dependent coordinates in the leg using acceleration level loop-closure constraints \cite{featherstone2014rigid}.
In this way, the equation of motion is written in terms of only the independent coordinates so that $\mathbf{q}=[\mathbf{q_{base}} \in \mathbf{SE}(3), \mathbf{q_{joints}} \in \mathbb{R}^{10}]$ and, 
\begin{align}
    &\mathbf{M(q)\Ddot{q} + C(q,\Dot{q})\Dot{q} + G(q)=B\tau + J^T F_{feet}}, \label{eq:3}
\end{align}
where $\mathbf{M(q)} \in \mathbb{R}^{n \times n}$ is the inertia matrix, $\mathbf{C(q,\Dot{q})\Dot{q}} \in \mathbb{R}^{n}$ includes Coriolis and centripetal forces, $\mathbf{G(q)} \in \mathbb{R}^{n}$ is the gravity term, and $\mathbf{B} \in \mathbb{R}^{n \times 10}$ is the actuator selector matrix given the torque at each actuated joint $\tau \in \mathbb{R} ^ {10}$.
The Jacobian to the feet is $\mathbf{J}$ and the ground reaction force applied to the feet is $\mathbf{F_{feet}}$.
The translational positions, velocities, and accelerations of the feet are a function of the generalized coordinates,
\vspace{-5pt}
\begin{align}
    \mathbf{x} &= \mathbf{FK(q)} \nonumber\\
    \mathbf{\dot{x}} &= \mathbf{J \dot{q}} \nonumber \\
    \mathbf{\ddot{x}} &= \mathbf{J \ddot{q} + \dot{J} \dot{q}}, \nonumber
\end{align}
\vspace{-5pt}
where $\mathbf{FK(q)}$ is forward kinematics of the legs.

\subsection{Task Space Control}
The task space inverse dynamics controllers consist of swing phase, stance phase, and the transition between the two. 
This basic setup allows the learning to explore in terms of different subfunctions in locomotion. $\mathbf{x}, \mathbf{x^{\delta}}, \mathbf{x^{ref}}$ are feet positions from the state estimator, the learned policy, and references. 

\subsubsection{Swing leg}
Swing leg control is formed as a PD acceleration controller with inverse dynamics on feet to reach to desired setpoints relative to base. First, the desired feet accelerations based on task space PD feedback is calculated as,
\begin{equation}
    \mathbf{\Ddot{x}^{des} = K_P^{swing}(x^{ref} + x^{\delta} - x) + K_D^{swing}(0 - \dot{x})}.  \nonumber
\end{equation}
Then we can compute the motor torques required to create the commanded task space acceleration using task space inertia, ($\mathbf{\Lambda = (JM^{-1}J^{T})^{-1}}$). 
We dropped the velocity product terms from \eqref{eq:3} and $\mathbf{\dot{J} \dot{q}}$ resulting in, 
\begin{equation}
    \mathbf{\tau_{swing} = J^T (\Lambda\Ddot{x}^{des} + \Lambda JM^{-1}G)}. \nonumber
\end{equation}

\subsubsection{Stance Leg}
The stance leg is primarily concerned with force interactions, so the control converts the set point commands into forces though the impedance controller with feedforward forces, 
\begin{equation}
    \mathbf{F^{des}_{feet} = K_P^{stance}(x^{ref} + x^{\delta} - x) + K_D^{stance}(0- \dot{x}) + F^{ref}}. \nonumber
\end{equation}
This force command is translated into joint torques through the simple Jacobian transpose control of feet to robot base, 
\begin{equation}
    \mathbf{\tau_{stance} = J^TF^{des}_{feet}}. \nonumber
\end{equation}
This is the torque that would be required to statically apply the force command in the absence of all other forces.

\subsubsection{Transitions between Stance/Swing}
The control structure also includes a transitioning rule to switch between the two phases of the gait, and relies on RL to find solutions to stabilize the robot. For each leg, the overall torque computation is, 
\begin{equation}
    \mathbf{\tau_{ID} = \phi \tau_{stance} + (\text{1}-\phi) \tau_{swing}},
\end{equation}
where $\phi \in [0,1]$ is a scalar output that depends on the phase along the target force profile on each leg, meaning if the target force profile is active, then $\phi = 1$ and if it is not active, then $\phi = 0$. This also allows for a double support phase where $\phi$ smoothly transits between stance and swing phases. During the transition phases around toe-off and touchdown, $\phi$ linearly interpolates between 0 and 1 and vice versa.

\subsubsection{Feet Orientation Control}
Swing and stance control only commands the translational positions of the feet. 
Foot orientation can be controlled via joint control of the hip yaw and foot pitch motors on each leg $\mathbf{\theta = \{\theta_{hipyaw}, \theta_{footpitch}\}}$. 
We use joint PD controllers on only the hip yaw and foot pitch motors. The policy outputs $\mathbf{\theta^{\delta}}$ to control the feet orientation to be flat and facing forward. $\mathbf{\theta^{ref}}$ represents a neutral joint position and is constant for each motor. 
These complimentary torques are defined as,  
\begin{equation}
\begin{aligned}
    \mathbf{\tau_{orient}} &=  \mathbf{\{\tau_{hip yaw}, \tau_{foot pitch}\}} \\
    &=\mathbf{ K_P^{joint}(\theta^{ref} + \theta^{\delta} - \theta) - K_D^{joint} \dot{\theta}}. \label{eq:foot}
\end{aligned}
\end{equation}


\subsubsection{Joint Level Commands and Control Rate}
The final torque commands to low-level actuators are, 
\begin{equation}
    \mathbf{\tau_{cmd} = \tau_{ID} + \tau_{orient}}. \label{eq:10} 
\end{equation}

As shown in Fig. \ref{fig:framework}, the full inverse dynamics control runs at 2kHz to compute torques based on \eqref{eq:10}. The RL policy outputs residual setpoints at 40Hz, meaning the desired target setpoints are updated at 40Hz.

\section{RL Problem Formulation}

\subsection{Learning Algorithm}
We use the Proximal Policy Optimization (PPO) algorithm \cite{Schulman2017}, a model-free policy gradient method with an actor-critic network. 
Both the actor and the critic use fully connected feed-forward neural networks with one hidden layer of size 256 and ReLU activation. 
Output layers use $tanh$ so that the policy outputs are bounded. 
We used the clipped objective for PPO with hyperparameters listed in the Appendix.

\subsection{Policy Inputs}
The input to the neural network policy consists of the observed robot state and user commands.
The entire observations $\mathbf{S \in \mathbb{R}}^{47}$ input to the neural network are measured through the proprietary state-estimator from Agility robotics. 
Using estimated states as inputs allows sim-to-real transfer without extra steps when considering possible differences between simulator and hardware.
The state space covers primarily the observations on the robot base and feet. 
More specifically, $\mathbf{S}$ includes commanded forward velocity $\mathbb{R}^{1}$, base's translational, rotational velocities $\mathbb{R}^{6}$ and orientations in quaternion $\mathbb{R}^{4}$, actuated and un-actuated joint positions and velocities measurable on hardware $\mathbb{R}^{28}$, and feet Cartesian positions relative to robot base $\mathbb{R}^{6}$. 
Also, $\mathbf S$ has an additional $\mathbb{R}^{2}$ input based on the phase $\phi$ variable along the references. 
This input is calculated with both sine and cosine functions to create unique phase identifiers and to wrap around the cycle. 

\subsection{Action Space}
The action space $\mathbf{A \in \mathbb{R}}^{10}$ is defined as residual terms, including the Cartesian positions of feet $\mathbb{R}^{6}$ relative to the robot floating-base and the joint positions of hip yaw and foot pitch $\mathbb{R}^{4}$ for foot orientation control in \eqref{eq:foot}. 

\subsection{Reward}
The reward centers on achieving task space control interests from the references for the robot base and end-effector feet. 
The desired reference quantities are defined with the corresponding phase along the references. We chose to not fully track the references because a tracking-only reward could result in the policy fitting too much to trajectories that do not have the notion of the full-order robot, making performance very brittle. 
Rather, we define a more general reward based on the fundamentals of locomotion: transitions between stance and swing. By focusing on task space costs this structure is flexible and can be applied to bipedal locomotion in general as opposed to constraining all joints to specific motion dictated by a robot specific trajectory. We still use the information encoded by the references by having them define the contact sequence and timing, but this method allows for more adaptive solutions. 


Rewards on the feet $r_{\text{lfoot}} \text{ or } r_{\text{rfoot}}$ are defined as locomotion subfunctions, swing and stance for each foot. 
Swing reward $r_{\text{foot-swing}}$ calculates the foot clearance cost weighted by swing foot velocities. Stance reward $r_{\text{foot-stance}}$ checks whether the foot is in contact with the ground by calculating both foot height and velocities. 
With the references from reduced-order models, the timing of swing and stance is predefined, as is the desired foot clearance height $h_{\text{swing-height}}$ which is set to 0.15m. 
Similar to transitions defined in controller \eqref{eq:10}, a continuous transition weighting $\phi \in [0, 1]$ indicates whether each foot should be in stance or swing,  
\begin{align}
     r_{\text{lfoot}} \text{ or } r_{\text{rfoot}} &= \phi r_{\text{foot-stance}} + (1-\phi) r_{\text{foot-swing}}.  \nonumber 
\end{align}
The reward function then consists of all the control interests on the robot floating-base defined by references on $r_{\text{base-xvel}}$,  $r_{\text{base-zvel}}$, and $r_{\text{base-zpos}}$, along with extra terms to encourage smooth, stable behaviors as,  
\begin{align}
    \text{R} &= 0.1 r_{\text{lfoot}} + 0.1 r_{\text{rfoot}} + 0.15 r_{\text{base-xvel}} + 0.1 r_{\text{base-zvel}} \\
    &+ 0.1 r_{\text{base-zpos}} + 0.1r_{\text{base-orientation}} + 0.15r_{\text{base-straight}} \nonumber \\ 
      & + 0.05r_{\text{foot-orientation}} + 0.025r_{\text{base-linear-accel}} \nonumber \\ 
      &+0.025r_{\text{base-angular-vel}} + 0.1r_{\text{action-smooth}} \nonumber 
\end{align}

Each reward term is a function of one or more physical quantities of the robot $s$ as described in Table \ref{tab:rewards}. The values are normalized using the exponential $r(s,a) =  \exp\Big(\frac{1}{T} \sum^{T}_{i=0}f(s,a)\Big)$ as a kernel function, creating a smooth reward between $(0, 1]$ for each reward term. Each $f$ is averaged over the duration $T=50$ between each policy update, given torque control frequency. The neutral orientation in quaternion $\mathbf{q_{\text{neutral}}}=(1,0,0,0)$.
\renewcommand{\arraystretch}{1.3}
\begin{table}[h!]
    \centering
    \vspace{-.25cm}
    \begin{tabular}{l | l}
      \hline			
      Terms ($f(s,a)$) & Expressions  \\
      \hline
      foot-stance & $x_{\text{foot-height}}^2 +  \|\mathbf{\Dot{x}_{\text{foot}}}\|_{2}$ \\
      foot-swing & $40 (h_{\text{foot-height}} - x_{\text{z-height}})^2$ \\
      
      base-xvel & $3|\Dot{x}_{\text{base}} - \Dot{x}_{\text{base-ref}}|$ \\
      
      base-zvel & $3|\Dot{z}_{\text{base}} - \Dot{z}_{\text{base-ref}}|$ \\
      
      base-zpos & $3|z_{\text{base}} - z_{\text{base-ref}}|$ \\
      
      base-orientation & $50(1-\langle \mathbf{q_{\text{base}}}, \mathbf{q_{\text{neutral}}} \rangle)$  \\
      
      base-straight & $5|y_{\text{base}}| + 3|\Dot{y}_{\text{base}}|$  \\
      
      foot-orientation & $30(1-\langle \mathbf{q_{\text{foot}}, q_{\text{neutral}}} \rangle)$  \\
      
      base-linear-accel & $\|\mathbf{\Ddot{x}_{\text{base}}}\|_{2}$
      \\
      
      base-angular-vel & $\|\bm{\omega_{\text{base}}}\|_{2}$ \\

      action-smooth & $3 \|\mathbf{a_{i} - a_{i-1}}\|_{2}$ \\
      \hline 
    \end{tabular}
    \caption{Reward Function Details}
    \label{tab:rewards}
    \vspace{-.4cm}
\end{table}
\section{Results}

\subsection{Experiment Setup}
\begin{figure}
    \centering
    \vspace{0.2cm}
    \includegraphics[width=1\columnwidth]{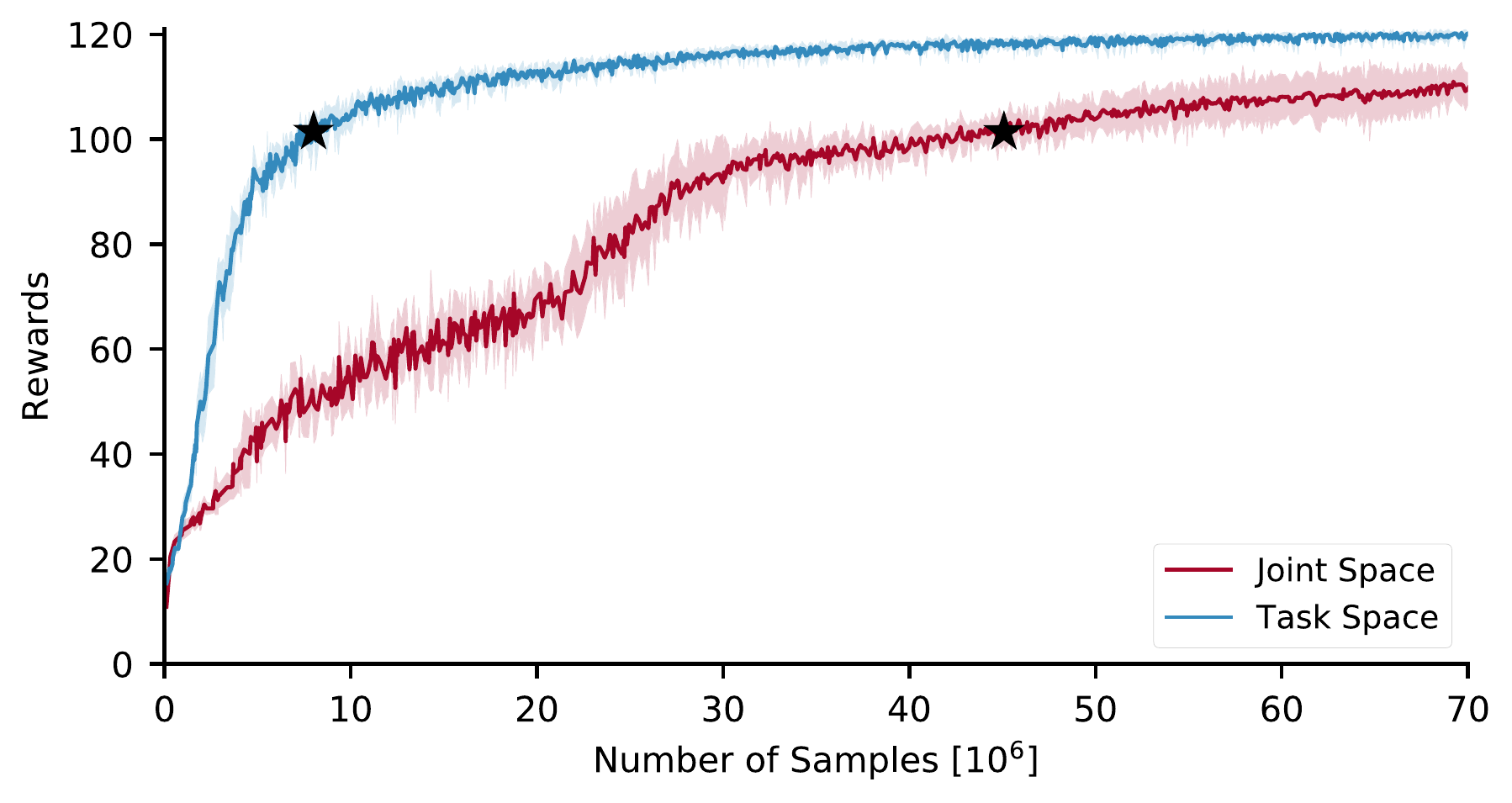}
    \caption{The learning curves show the average of five random seeds for each action space. Actions in task space are able to learn rough stepping gaits at the beginning of the training and quickly converge to a plateau. However, actions in joint space need to explore for joint coupling before the agents can effectively explore in a more useful episode. Star represents when we observe the robot starting to walk stably in simulation with the learned policy, which takes around 8 and 32 hours (wall-clock time) for task space and joint space respectively. }
    \label{fig:sample}
\end{figure}


\begin{figure*}[ht!]
    \centering
    \vspace{0.3cm}
    \includegraphics[width=0.9\linewidth]{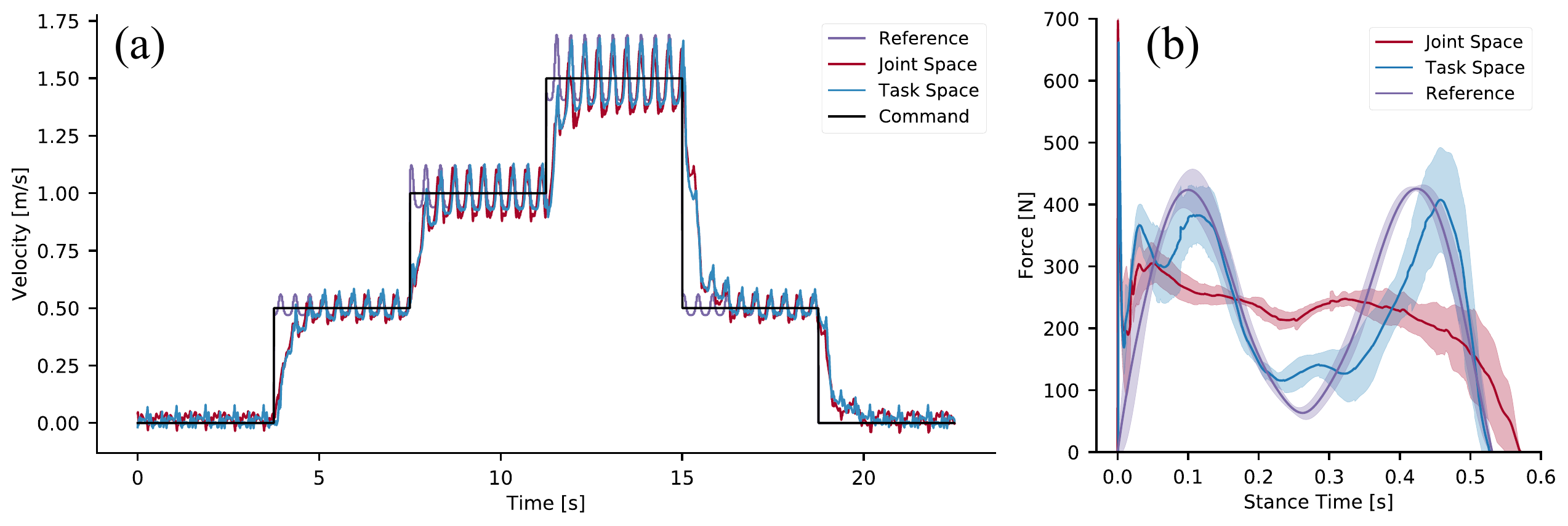}
    \caption{Simulation results for learned policies. Plot (a) shows a speed tracking comparison. Plot (b) shows the ground reaction forces recorded from the speed tracking test averaged over the stance period.}
    \label{fig:sim}
    \vspace{-.5cm}
\end{figure*}

For our experiments, the learning problem is given with a set of reference signals $\mathbf{x^{ref}, F^{ref}}$ that are optimized based on the actuated spring mass models in the extended work from \cite{Green2020}. 
These trajectories are not dynamically feasible because of the simplified model, but are useful feedforward signals for the inverse dynamics controller.
The gains for the task space controllers are $\mathbf{K_P^{swing}}=\{300, 150, 400\}\text{s}^{-2}$ and $\mathbf{K_D^{swing}}= \{10, 3, 10\}\text{s}^{-1}$ for the forward/lateral/vertical directions respectively. $\mathbf{K_P^{stance}}$ and $\mathbf{K_D^{stance}}$ have the same corresponding values but with units N/m and Ns/m.
The gains for the hip yaw motors and the foot pitch motors are $\mathbf{K_P^{joint}}=$\{100, 50\}$\text{Nm/rad}$ and $\mathbf{K_D^{joint}}=$\{10, 5\}$\text{Nm/rads}^{-1}$ respectively.
The gains are the same in simulation and on hardware, and are tuned to track task space targets without stability problems across different configurations.

To compare the proposed method with actions learned in joint space on the sample efficiency and task space dynamics, we implemented the joint space action learning approach where policy outputs the residual joint setpoints with gains and PD controller similar to \cite{Xie2019}. 
We used the same references signals from the spring mass model for both action spaces, allowing both action spaces to use the same references for reward and control. The joint space references are numerically derived using offline inverse kinematics on the same set of reference motions.

For the sampling and training process, we use \emph{cassie-mujoco-sim} simulation packages\footnote{https://github.com/osudrl/cassie-mujoco-sim} based on the Mujoco physics simulator \cite{Todorov2012}. 
The training horizon for each episode is 150 steps, which corresponds to running the policy at 40Hz for 3.75 seconds. We sample 5000 simulation timesteps for each iteration of training.
During the learning process, the initial conditions of the robot are randomized for each episode, including the reference phase and the velocities of the robot base to simulate disturbances. 
The termination condition of each episode only considers if the robot base height falls below 0.6 meters.
Policies were trained using 50 cores on a dual Intel
Xeon
Platinum 8280 server.

\subsection{Simulation Results}
We primarily focused on comparisons in simulations for sample efficiency and how well each action space achieves the desired task space control interests. 

\subsubsection{Sample Efficiency}
We trained policies for both action spaces to compare sample efficiency under the same reward and learning setup. 
Fig. \ref{fig:sample} shows that learning task space actions reaches high reward much faster than learning joint space actions. 
This increase in sample efficiency has a large practical impact, decreasing wall-clock training time from 32 hours to 8 hours in order to obtain a policy that starts to walk stably.
During the early stages of training, actions in task space have a direct effect on the robot dynamics by moving the feet around in coordination, as opposed to actions in joint space which results in the robot standing still without any intentions to move the robot base towards the desired velocities. We show the during-training comparison in the video and discuss the potential cause further in Section \ref{sec:explore_noise}.

\subsubsection{Commanded Velocity Tracking}
We tested the steady-state locomotion by sending varying desired speed commands into the corresponding input to the policy. During training, desired speed for each episode holds constant, but the robot is initialized with random velocity of the robot base. The final learned policy can interpolate across the desired speed range as shown in Fig. \ref{fig:sim}(a). Both action spaces have comparable performance on speed tracking. 

\subsubsection{Ground Reaction Forces}
We used the optimized reduced-order references to derive the references in task space and joint space. 
By following the desired CoM dynamics defined in the reward, the ground reaction forces should capture the trend described from the references as seen from Fig. \ref{fig:sim}(b). 
Ground reaction forces are recorded during the speed tracking experiments where Cassie walks across the speed range. 
Although policies by both action spaces are able to produce stable locomotion behaviors, the underlying ground reaction forces are very different. 
Directly learning task space actions produces the desired ground reaction forces over repeated gait cycles, whereas joint space actions cannot reproduce the desired stance dynamics even after reward plateaus. 
Furthermore, we can see that the task space actions are not exactly following the feedforward force profile, but rather adapting the profile and stretching the mid to late stance to achieve better stability.  


\subsection{Exploration Noise in Action Space}
\label{sec:explore_noise}
\begin{figure}
    \centering
    \includegraphics[width=0.9\columnwidth]{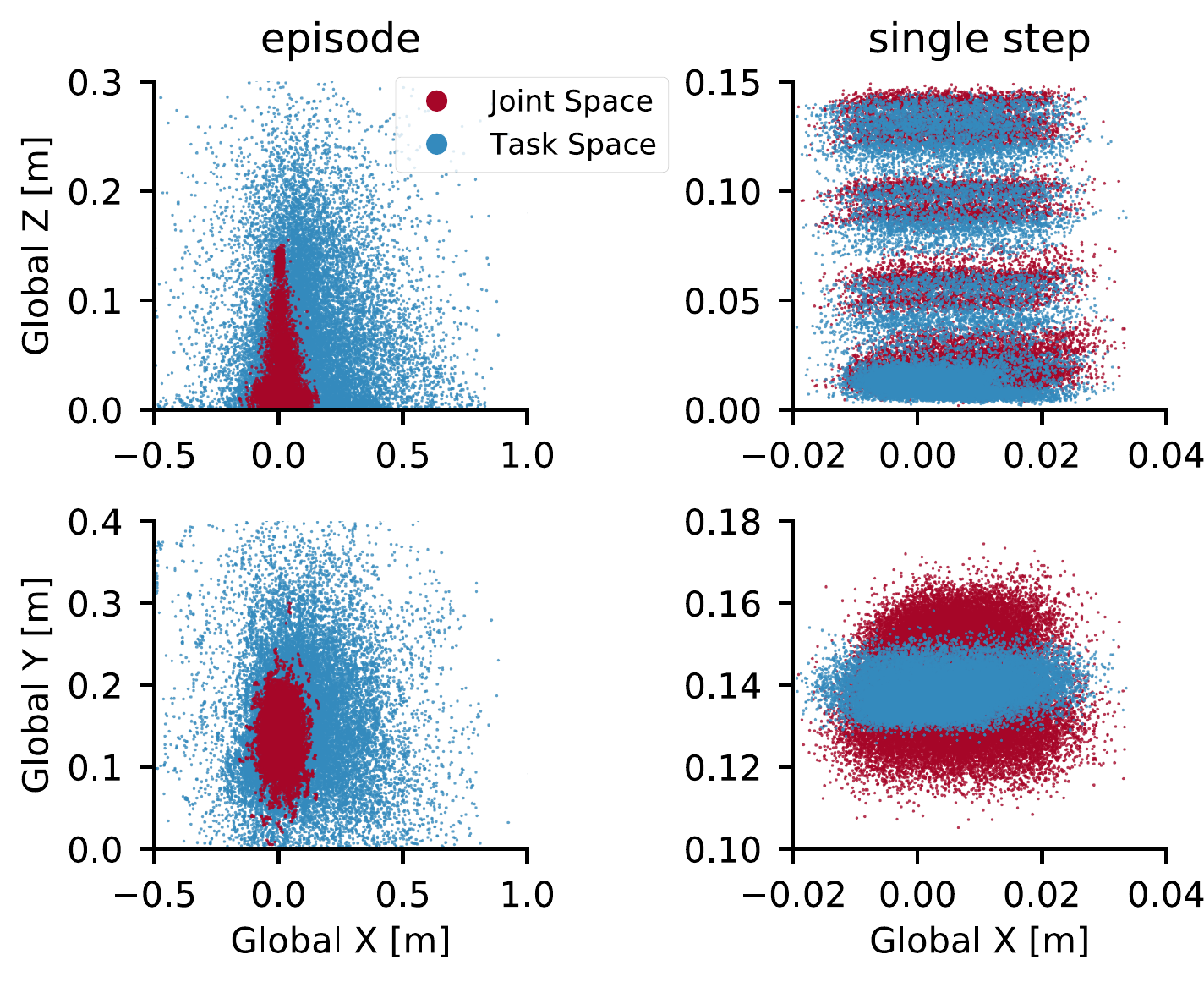}
    \caption{
    Distributions of foot locations under random action noises for each action space. Single step shows both action spaces covering a similar space of foot locations. Task space shows a more direct exploration over the entire episode that results into a larger space covered of feet movements compared to joint space. 
    ``Stripe-like" samples are a result of using a small set of initialization states of the robot.
    }
    \label{fig:noise}
\end{figure}

Noise in learning algorithms allows agents to explore their environment but it can make learning very brittle without the proper tuning.
For legged locomotion tasks, the robot dynamics are primarily influenced through the motion of the feet. In order to evaluate the effective exploration by both action spaces, we compare the size of the space covered by the feet during random exploration. 
In our experiments we find that the exploration noise $\sigma$ in the Gaussian policies must be different for each action space in order to produce feasible policies. 
Thus, we analyzed the effects of the noise for each action space independent of the particular learning algorithms.  
Fig. \ref{fig:noise} shows the distribution comparison of sampled task space locations when sampling in each action space. 
Actions are drawn from a Gaussian distribution $a \sim \mathcal{N}(0,\,\sigma^{2})$, where $log\sigma = -1.5$ for joint space and $log\sigma = -2.5$ for task space, same as used in PPO in the Appendix. 
First, we collected data where we apply the action noise to a single 40Hz timestep.
This allows us to compare the joint space and task space action distributions.
In the single step plots on the right we can see that the distributions are comparable, with joint space exploring slightly more in the Y (lateral) direction.

We then test how the action noise propagates through an entire episode, where random actions are continually drawn and acted until a termination condition is satisfied or the horizon is reached. We collected 50,000 samples for each scenario.
We see that over a whole episode that the task space sampling covers a much larger area of the state space in the episode scenario. Furthermore, since termination conditions are used, the data that task space sampling generates captures dynamics before falling. In other words, it is not just the case that task space sampling generates a larger range of states, it generates a larger range of \textit{useful} states. We noticed that simply increasing the noise in joint space to $log\sigma = -1$ during sampling did not increase the feet range of motions, because the extra noise would just cause the robot to fall down and lead to bad states cutoff by the termination condition. This result highlights the need for effective \textit{directed} exploration for legged locomotion in its natural action space. One cannot just use more random noises in an effort to sample more states to improve learning. Task space both contains the primary control interests and allows for directed exploration that greatly improves sample efficiency. 

\subsection{Hardware Transfer}
We are able to transfer the learned policies onto Cassie without additional randomizations on robot model parameters. 
The supplementary video shows the robot is capable of stepping in-place and walking at various speeds on treadmill.
\section{CONCLUSION} 

In this work, we present an approach that combines bipedal robot control with reinforcement learning to enable a policy to learn actions in task space. 
We show that learning actions in task space can produce steady-state locomotion behaviors and demonstrated the successful sim-to-real transfer on Cassie. 
Specifically, the learned actions represent setpoint residuals relative to the robot base. 
By varying this setpoint over the stance and swing phases, the learned policies adapt the references trajectories while being able to reject disturbances.

The proposed approach unloads RL from having to re-learn the known robot models, so it is sample efficient and explores the task space regions more broadly to find solutions. We found that simply increasing the action noises in joint space does not correlate to better exploration, because undirected exploration in Gaussian policies may gather many useless samples. 
On the other hand, directly giving such joint coordination information into learning may hinder the discovery of interesting coupling effects among the joints. 
Further research may focus on how to embed the notions of the effective task space changes into the learning process.
Learning task space action could also allow RL to communicate in the same language with potentially various layers of a control hierarchy, where exchange signals are mostly defined in task space.
Future research may benefit on incorporating more advanced model-based control designs, such as adding centroidal dynamics as one of the actions.

\begin{appendix}
Hyperparameters for Proximal Policy Gradient,
\renewcommand{\arraystretch}{1.2}
\begin{table}[h!]
    \centering
    \vspace{-.25cm}
    \begin{tabular}{l | l}
      \hline			
      Parameter & Value  \\
      \hline
      Horizon ($H$) & 150 \\
      Adam stepsize & 1$\times10^{-4}$  \\
      Number epochs & 8 \\
      Minibatch size & 1024  \\
      Discount ($\gamma$) & 0.99\\
      Clipping parameter ($\epsilon$) & 0.2 \\
      Max gradient norm & 0.05\\
      Log stddev. of action distribution & -2.5 (task space) \\
      & -1.5 (joint space) \\
      \hline 
    \end{tabular}
    \label{tab:ppo_hyper}
    \vspace{-.4cm}
\end{table}

\end{appendix}


\section*{Acknowledgments}
\small{We thank Intel for providing vLab resources. We also thank DRL members and Jianwen Luo for helpful discussions.}

\def\bibfont{\footnotesize}
\bibliographystyle{IEEEtranN}
\bibliography{main}

\end{document}